\documentclass[11pt,a4paper]{article}
\usepackage[hyperref]{emnlp-ijcnlp-2019}
\usepackage{times}
\usepackage{latexsym}

\usepackage{amsfonts}
\usepackage{url}
\usepackage{xcolor}
\usepackage[ruled,linesnumbered]{algorithm2e}

\SetCommentSty{mycommfont}
\usepackage{graphicx}
\usepackage{bbm}

\usepackage{amsmath}
\DeclareMathOperator*{\argmax}{arg\,max}

\aclfinalcopy 

\title{A Failure of Aspect Sentiment Classifiers and \\ an Adaptive Re-weighting Solution}

\author{Hu Xu\textsuperscript{\text{1}}, Bing Liu\textsuperscript{\text{1}}, Lei Shu\textsuperscript{\text{1}}\and Philip S. Yu\textsuperscript{\text{1,2}}\\
    \textsuperscript{1}{Department of Computer Science, University of Illinois at Chicago, Chicago, IL, USA}\\
    \textsuperscript{2}{Institute for Data Science, Tsinghua University, Beijing, China}\\
    {\tt \{hxu48, liub, lshu3, psyu\}@uic.edu}
}

\date{}

\begin{document}

\maketitle

\begin{abstract}
\underline{A}spect-based \underline{s}entiment \underline{c}lassification (ASC) is an important task in fine-grained sentiment analysis.~Deep supervised ASC approaches typically model this task as a pair-wise classification task that takes an aspect and a sentence containing the aspect and outputs the polarity of the aspect in that sentence. However, we discovered that many existing approaches fail to learn an effective ASC classifier but more like a sentence-level sentiment classifier because they have difficulty to handle sentences with different polarities for different aspects.~This paper first demonstrates this problem using several state-of-the-art ASC models. It then proposes a novel and general \underline{a}daptive \underline{r}e-\underline{w}eighting (ARW) scheme to adjust the training to dramatically improve ASC for such complex sentences. Experimental results show that the proposed framework is effective \footnote{The dataset and code are available at \url{https://github.com/howardhsu/ASC_failure}.}.

\end{abstract}

\section{Introduction}
\label{sec:intro}

\underline{A}spect-based \underline{s}entiment \underline{c}lassification (ASC) is an important task in fine-grained sentiment analysis~\cite{hu2004mining,liu2015sentiment}, which aims to detect the opinion expressed about an aspect (of an opinion target).
It not only requires fine-grained annotation of aspects and their associated opinions, but also more sophisticated classification methods.
Unlike document-level sentiment classification where opinion terms appear frequently in a document, so it is easier to detect the overall sentiment/opinion of the document \cite{pang2002thumbs,liu2015sentiment}, detecting aspect-level sentiments in short text (e.g., a sentence) requires more accurate understanding of very fine-grained opinion expressions and also correct association of them with each opinion target (or aspect).
For example, ``The screen is good but not the battery'' requires to detect two fine-grained and \textit{contrastive opinions} within the same sentence: a positive opinion towards ``screen'' and a negative opinion towards ``battery''. We found that existing ASC models have great difficulty to correctly classify such contrastive opinions in their sentences. 

\begin{table}
    \centering
    \scalebox{0.7}{
        \begin{tabular}{l|c|c}
            \hline
            Review Sentence & Sent.-level & Asp.-level \\
            \hline
            The screen is good. & pos & screen: pos \\
            \hline
            The screen is good and also the battery. & pos & screen: pos \\
            & & battery: pos \\
            \hline
            The screen is good but not the battery. & contrastive & screen: pos\\
             &  & battery: neg\\
            \hline
        \end{tabular}
    }
    \caption{A few sentences for ASC with both sentence-level(sent.-level) polarity and aspect-level(asp.-level) polarity: the first two sentences can leverage sentence-level polarity to answer aspect-level polarity correctly but not for the last (contrastive) sentence.}
    \label{tbl:ex_failure}
\end{table}

Deep supervised ASC approaches typically model ASC as a memory network~\cite{weston2014memory,sukhbaatar2015end,tang2016aspect}. Given two inputs: a sentence $x$ and an aspect term $a$ appearing in $x$, build a model $p_\theta(\hat{y}|a, x)$, where $\hat{y} \in \{\textit{pos}, \textit{neg}, \textit{neutral}\}$ is the opinion (or sentiment) about $a$.
From the perspective of classification, this formulation is essentially a pair-wise classification task that takes a pair of inputs $(x_j, x_k)$ and predicts the class $p_\theta(\hat{y}|x_j, x_k)$. 
One challenge of pair-wise classification is the quadratic space of combinations introduced by the two inputs.
This requires a huge number of critical training examples to inform the model what the learning task is and what kinds of interactions between the two inputs are necessary for that task.

For ASC, we discovered that the available datasets may not provide such rich interactions for effective supervision. 
In fact, we observed that lacking of sentences with contrastive opinions (we call it \textbf{contrastive sentence} for brevity) can make an ASC classifier downgrading to a \underline{s}entence-level \underline{s}entiment \underline{c}lassifier (SSC), as intuitively explained in Table \ref{tbl:ex_failure}.
By ``contrastive'', we mean two or more different opinions are associated with different aspects appearing in the same sentence.
After all, when showing training examples with only sentences of one or more aspects with the same opinion (or polarity), the pair-wise model (or humans) can totally ignore the aspect part $a$ and only use the sentence $x$ to classify the aspect-level opinion correctly with an overall sentence-level opinion.
Contrastive sentences are crucial for ASC, but they are infrequent, as we will see in the Dataset Analysis section.
As a result, contrastive sentences are largely ignored in training and further weakly evaluated in the testing. 
This results in the failure of the current ASC models in correctly classifying contrastive opinions, as shown in the Experiments section.
In fact, this is a general issue for most machine learning models, where the majority wins and dominates the training process and the rare but important examples can easily be ignored and may even be considered as noise, as seen in many class imbalance problems and machine learning fairness problems. 
For example, the object detection problem \cite{shrivastava2016training,lin2017focal} in computer vision can easily come up with long-tailed and imbalanced classes of examples given it is almost impossible to manually rebalance objects appear within an image.

In this paper, we assume that available datasets can be easily and unintentionally imbalanced by following the distributions naturally in reviews.
We propose to apply a weight to each training example, representing the importance of such an example during training.
We investigate different methods of computing weights and propose a training scheme called \underline{a}daptive \underline{r}e-\underline{w}eighting, which dynamically keeps the system focusing on examples from contrastive sentences.
We show that 
a model trained with such a scheme can improve the classification of examples from contrastive sentences dramatically, while still keep competitive or even better performance on the full set of testing examples. 

The main contribution is 2-fold: 
(1) It discovers the issue of ASC that plagues existing methods, which are clearly manifested in contrastive sentences. Such sentences are essential for the ASC task but are largely ignored.
(2) It proposes a re-weighting solution that resolves the issue and improves the performance on both contrastive sentences and the full set of testing examples.


\section{Dataset Analysis}
\label{sec:da}

We adopt the popular SemEval 2014 Task 4\footnote{http://alt.qcri.org/semeval2014/task4} datasets to demonstrate how rare those contrastive sentences are. These datasets cover two domains: \emph{laptop} and \emph{restaurant}. We further demonstrate that the normal training on such datasets results in poor performances on contrastive sentences in experiments.

\begin{table}[t]
\centering
\scalebox{1.0}{
    \begin{tabular}{l|c|c}
    \hline
    & \bf{Laptop} & \bf{Restaurant} \\
    \hline
    \bf{Training} & & \\
    \hline
    \#Sentence & 3045 & 2000 \\
    \#Aspect & 2358 & 1743 \\
    \#Positive & 987 & 2164 \\
    \#Negative & 866 & 805 \\
    \#Neutral & 460 & 633 \\
    \#Sent. /w Asp. & 1462 & 1978 \\
    \hline
    \#Contrastive Sent. & \textbf{165} & \textbf{319} \\
    \%Contrastive Sent. & \textbf{11.3\%} & \textbf{16.1\%} \\
    \hline
    \hline
    \bf{Testing Set} & & \\
    \hline
    \#Sentence & 800 & 676 \\
    \#Aspect & 654 & 622 \\
    \#Positive & 341 & 728 \\
    \#Negative & 128 & 196 \\
    \#Neutral & 169 & 196 \\
    \#Sent. /w Asp. & 411 & 600 \\
    \hline
    \#Contrastive Sent. & \textbf{38} & \textbf{80} \\
    \%Contrastive Sent. & \textbf{9.2\%} & \textbf{13.3\%} \\
    \hline
    
    \hline
    \end{tabular}
}
\caption{Summary of SemEval14 Task4 on aspect sentiment classification. \#Sentence: number of sentences; \#Aspect: number of aspects; \#Positive, \#Negative, and \#Neutral: number of aspects with positive, negative and neutral opinions, respectively; \#Sent. /w Asp.: number of sentences with at least one aspect that is associated with one of positive, negative or neutral opinion; \#Contrastive Sent.: number of sentences with aspects associated with different opinions; \%Contrastive Sent.: percentage of contrastive sentences in sentences with at least one aspect.}
\label{tbl:asc}
\end{table}

As shown in Table \ref{tbl:asc}, we first examine the overall statistics of these datasets.
We decompose these statistics to get deeper insights that may lead to a failed ASC classifier. 
We notice that although these datasets contain a moderate number of training sentences for laptop, sentences with at least one aspect (and thus has polarities of opinions) is less than 50\%, as in \emph{\#Sent. /w Asp}. 

Further, as discussed in the introduction, we are particularly interested in contrastive sentences that have more than one aspect and are associated with different opinions (\textit{\#Contrastive Sent.}) for each such sentence.
Those sentences carry critical training examples (information) for ASC because the rest of the examples have only one polarity per sentence (even with two or more aspects), where the overall sentence-level opinion can be applied to aspect-level opinion and thus effectively downgrade the task of ASC to SSC (sentence-level sentiment classification).

We notice that contrastive sentences are rare in both training and test sets of both domains. Laptop is even more so on the shortage of contrastive sentences because of the shortage of sentences with at least one aspect.
If we consider their percentage (\textit{\%Constrastive Sent.}), the training set of restaurant has just about \textbf{16\%} and the laptop has only about \textbf{11\%}. With the SSC-like examples dominating the training set, a machine learning model trained on such a set is susceptible to ignoring the aspect and mostly performing SSC.

\begin{table}[t]
\centering
\scalebox{1.0}{
    \begin{tabular}{l|c|c}
    \hline
    & \bf{Laptop} & \bf{Restaurant} \\
    \hline
    \bf{Contrastive Test Set} & & \\
    \hline
    \#Contrastive Sent. & \textbf{78} & \textbf{80} \\
    \#Aspect & 203 & 228 \\
    \#Positive & 72 & 85 \\
    \#Negative & 71 & 60 \\
    \#Neutral & 60 & 83 \\
    \hline
    \end{tabular}
}
\caption{Summary of Contrastive Test Set.}
\label{tbl:our}
\end{table}

What is worse is that the test set for the \emph{laptop} domain contains only 38 contrastive sentences.
This further poses a challenge on evaluating the ASC capability for \emph{laptop} as only contrastive sentences can evaluate the true capability of ASC. 
To solve this problem, two (2) annotators are asked to follow the annotation instructions of SemEval 2014 Task 4 and annotate more contrastive sentences (to have a similar number of contrastive sentences as \emph{restaurant} in total) from Laptop reviews~\cite{he2016ups}.
Disagreements are discussed until agreements are reached.
The main complaint from the annotators is that finding such sentences takes a lot of time as they are infrequent.
By combining the additional contrastive sentences with those contrastive sentences from the original test set, we form a new \emph{contrastive test set}, dedicated to testing the true ASC capability of ASC classifiers. Note that there is no change to the training set for the laptop domain and no change to either the training or the test set of the restaurant domain. The final statistics of the contrastive sentence are shown in Table \ref{tbl:our}.
To simplify our description, we refer the original test set as the \emph{full test set}.
Note that we DO NOT add those extra contrastive sentences into the full test set to keep the results comparable with existing approaches.
We evaluate the failure of existing ASC classifiers on the contrastive test set in experiments and discuss our example re-weighting scheme that focuses training on rare contrastive sentences in the next section.


\section{Adaptive Re-Weighting}
\label{sec:arw}
In this section, we first describe the motivation for developing a new training scheme instead of following the canonical training process. Then we describe the general idea of designing the \underline{a}daptive \underline{r}e-\underline{w}eighting (ARW) scheme and the detailed scheme afterward. 

\subsection{Motivation}
Given that the examples from contrastive sentences are rare, the first question that one may ask is how a deep learning model learns from those rare examples during the existing training process.
Existing research showed that rare and noisy examples are seldom optimized at the early stage of training (e.g., a few epochs)~\cite{gao2016sample}.
Intuitively, in the beginning, the losses from the majority examples dominate the total loss, and they determine the direction of parameter updates based on their gradients. 
So the losses from majority examples can be smaller in the next few iterations. 
At a later stage, although the loss from a rare example can be larger than the one from a majority example, the rare example still may not have enough contribution to the total loss as the loss in each batch is averaged among all examples, although the losses from the majority examples are smaller now.
Also, as the rare examples can be rather diverse, it is unlikely that a similar rare example can later appear in another batch to have a similar impact.
In the worst case, it is possible that the rare examples' losses are taken care of when the optimizer starts to overfit the minor details in the majority examples. When the validation process kicks in for early stopping, which aims to avoid overfitting, it may stop training the model before rare examples are really optimized well.
To demonstrate our observation, we show how many incorrectly classified training examples are from contrastive sentences in experiments.

Given this unwanted behavior of optimization, a natural idea to solve the rare example problem is to detect those examples from contrastive sentences at an earlier stage of training and increase (or rebalance) their losses much earlier before the validation process finds the best model.
One natural solution to increase those losses is to give higher weights to those examples from contrastive sentences that are not optimized well.
Then the total loss (per batch of optimization) is the weighted sum of losses of examples (within a batch).
This process of adjusting example weights could be dynamic in nature because a used-to-be easy example can be an incorrect one later and vice versa.
For example, in ``The screen is good but not the battery.'', increasing the loss for aspect ``battery'' can make the ``screen'' incorrect later, leading to increase the loss for ``screen'' later.
Further, note that although the model can easily access contrastive sentences based on the polarity labels during preprocessing/training, the model has no access to which example is from a contrastive sentence during validation or testing. Tackling those sentences must be automatically done during training.

\begin{algorithm*}[!pt]
    \LinesNumbered
    \DontPrintSemicolon
    \caption{\underline{A}daptive \underline{R}e-\underline{w}eighting (ARW) Scheme}
    \label{alg:arw}
    \SetKwInOut{Input}{Input} 
    \SetKwInOut{Output}{Output} 
    \Input{$\mathcal{D}_{\text{tr}}$: training set with $n$ examples; \\$e$: maximum number of epochs.} 
    \Output{$p_\theta(\hat{y}|\cdot, \cdot)$: a trained model.}

    \BlankLine


    $w_{1:n} \gets \frac{1}{n} $ \tcp*{Initialize all example weights uniformly.}
    \For{$\text{epoch} \in \{1, \dots, e\}$ \tcp*{Pass through the training data epoch-by-epoch.} }{
        \For{$(a^b, x^b, y^b, w^b) \in \text{Batchify}(\mathcal{D}_{\text{tr}}, w_{1:n})$ \tcp*{Retrieve one randomly sampled batch.}}{
            $l^b \gets \text{CrossEntropy}(p_\theta(\hat{y}^b|a^b, x^b), y^b)$\tcp*{Compute example-wise loss.}
            $L^b \gets \frac{\sum(w^b \cdot l^b)} {\sum w^b}$ \tcp*{Re-normalize weighted loss and compute total loss.}
            $\text{BackProp\&ParamUpdate}(L, M)$   \tcp*{Back propagation and parameter updates.}
        }
        $\hat{y}_{1:n}\gets \argmax p_\theta(\hat{y}_{1:n}|a_{1:n},x_{1:n})$ \tcp*{Compute current prediction.}
        $r \gets \frac{\sum_{i=1}^{n}(w_i\mathbb{I}[y_{i}\neq \hat{y}_{i} \wedge \text{Contra}(x_i)])}{\sum_{i=1}^n w_i}$ \tcp*{Compute weighted error rate.}
        $\alpha \gets \log(\frac{(1-r)+\epsilon}{r-\epsilon})$ \tcp*{Compute the log correct-incorrect ratio.}
        $w_{1:n} \gets w_{1:n} \exp(\alpha \mathbb{I}[y_{1:n} \neq \hat{y}_{1:n}) \wedge \text{Contra}(x_{1:n}) ])$ \tcp*{Adjust weights of incorrect examples.}
        
    }
\end{algorithm*}

\subsection{Proposed Training Scheme}

Given the above analysis, we aim to design an adaptive scheme that keeps adjusting the weights for losses of examples from contrastive sentences (which are known in the training set).
Increasing losses can be modeled as having weights $w_{1:n}$ assigned to the $n$ training examples and the total loss $L$ is computed as the weighted sum of the training examples. 
So an example with a higher weight is more likely to contribute more to $L$.
As deep learning models are typically trained on a batch-by-batch basis, we define the total loss $L^b$ as the loss from a batch.
Let $l^b$ be the example-wise losses for examples within a batch. 
Since a batch is randomly drawn from the training set, we re-normalize the weights $w^b$ for examples in that batch $L^b=\frac{\sum(w^b \cdot l^b)} {\sum w^b}$ to avoid fluctuation caused by randomly drawing examples with weights of different magnitude.

Then the next issue is when and how to adjust the weights. 
We assume an uniform distribution for weights at the beginning $w_{1:n} \gets \frac{1}{n} $.
A natural point of adjusting the weight for each example is at the end of training of each epoch.
This is because every example has been consumed once and the model can focus on examples from contrastive sentences that are not treated well (incorrectly classified).
To adjust the weights, the first step is to find incorrect examples as $y_{i}\neq \hat{y}_{i}$ for $i \in [1, n]$, where $\hat{y}_{i}$ is the prediction of the $i$-th training example from the current model and $y_{i}$ is the ground-truth.
Then we pick those incorrect examples that are from contrastive sentences as an indicator variable $\mathbb{I}[y_{i} \neq \hat{y}_{i} \wedge \text{Contra}(x_{i})]$, where $\text{Contra}(\cdot)$ tells whether the sentence $x_i$ is a contrastive sentence or not.
$\text{Contra}(\cdot)$ requires preprocessing to know which sentence has more than one polarity of aspects (the training data contains the information).
For research question 5 (RQ5) in experiments, we perform an ablation study on whether this term is important.
Note that existing research (e.g., \cite{lin2017focal} in object detection) favors a continuous loss-based weighting function over the correctness-based weighting function.
We realize that correctness-based weighting function is better on which examples should improve when we address RQ4 in experiments.

Now we estimate the overall weighted error rate $r \in [0, 1]$ to detect whether the current model tends to make more mistakes on contrastive sentences or not. Note that the reason for using the weighted error rate instead of just the error rate is that the weighted error rate reflects the hardness on optimizing examples from contrastive sentences instead of simply example-level errors. We will detail the formula in the next subsection.
When the weighted error rate is high (e.g., $>0.5$), instead of increasing the weights for incorrect examples from contrastive sentences, we probably need to reduce them so as to avoid learning too much noise.
Lastly, the weight adjustment for incorrect examples from contrastive sentences is determined by the (correct-versus-incorrect) ratio $(\frac{(1-r)+\epsilon}{r-\epsilon})$. So when this amount is larger than $1$, multiplying it to increase the weights and otherwise to decrease the weights.
Here we introduce a weight assignment factor $\epsilon$, which is a hyperparameter to control whether the model should favor even more weights (e.g., $\epsilon>0$) or not (e.g., $\epsilon<0$).
We detail the proposed ARW algorithm in the next subsection.



\subsubsection{ARW Algorithm}
The proposed ARW algorithm is shown in Algorithm \ref{alg:arw}.
In Line 1, it initializes the weights of all training examples uniformly.
Lines 2-14 pass through the training data epoch-by-epoch and update the example weights at the end of each epoch.
Specifically, Line 3 retrieves one randomly sampled batch of aspects $a^b$, sentences $x^b$, polarity label $y^b$ and their (current) corresponding weights $w^b$.
Line 6 makes a forward pass on aspects and sentences $p_\theta(\hat{y}|a^b, x^b)$. Then we compute example-wise loss $l^b$ for each training example in the batch.
Line 7 computes the weighted loss and re-normalize these weights throughout the batch to get the total loss $L^b$. 
Line 8 does normal backpropagation and parameter updating as in ordinary neural networks training.
Line 10 gets the prediction on the training set.
Line 11 first discovers the hard examples represented by an indicator variable $\mathbb{I}[y_{i}\neq \hat{y}_{i} \wedge \text{Contra}(x_i) ] $.
It then computes the weighted error rate. 
Line 12 computes the log of the correct-incorrect ratio. $\alpha >0$ indicates increasing the weights and $\alpha<0$ means decreasing the weights.
Lastly, in Line 13, we only adjust the weights via the indicator variable $\mathbb{I}[y_{1:n} \neq \hat{y}_{1:n} \wedge \text{Contra}(x_{1:n})]$ since the weights of correctly classified (easy) examples are always multiply by $1$.
As a result, Algorithm \ref{alg:arw} keeps track of the weights $w_{1:n}$ for all training examples and always focuses on adjusting weights of incorrect examples from contrastive sentences. We also perform a normal validation process after each epoch (omitted in the Algorithm \ref{alg:arw} for brevity).

\section{Experiments}
\label{sec:exp}

Our experiment consists of two parts: (1) show the failure of existing approaches and (2) demonstrate the effectiveness of the ARW scheme.
We focus on the following research questions (RQs):\\
\textbf{RQ1}: How is the performance of existing ASC systems on the contrastive sentences in the test data (\emph{Contrastive Test Set}) ?\\
\textbf{RQ2}: What is the performance of an ASC model trained from data with manually assigned fixed higher weights to contrastive sentences only? \\
\textbf{RQ3}: How is the performance of the proposed ARW system compared with the above baselines?\\
\textbf{RQ4}: How is the performance of a loss-based weighting function (such as the famous focal loss \cite{lin2017focal}) compared to ARW?\\
\textbf{RQ5}: How important is the term $\text{Contra}(\cdot)$ (in Lines 11 and 13), given it needs preprocessing to find which sentence is contrastive?\\
\textbf{RQ6}: Can ARW tackle more examples from contrastive sentences before early stopping (via the validation set) ?

\subsection{Failure of Existing Approaches}
\subsubsection{ASC Baselines}
\label{sec:baselines}

To demonstrate existing ASC systems' difficulty with contrastive sentences, we used a range of ASC baselines and tested their performance on examples from contrastive sentences (contrastive test set). We evaluate all baselines on both accuracy (Acc.) and macro F1 (MF1).

\noindent
\textbf{RAM}\cite{chen2017recurrent}\footnote{The first 4 baselines are adopted from \url{https://github.com/songyouwei/ABSA-PyTorch}.}. 
This system proposes a multiple-attention mechanism to capture sentiment features separated by a long distance so that it is more robust against irrelevant information. The weighted-memory and attention mechanism not only helps avoid the labor-intensive feature engineering work but also provides a tailor-made memory for different opinion targets of a sentence.\\
\textbf{AOA}\cite{huang2018aspect}. This system introduces an attention-over-attention (AOA) neural network, which models aspects and sentences in a joint manner and explicitly captures the interaction between aspects and the sentence context.\\
\textbf{MGAN}\cite{li2018exploiting}. This method leverages the fine-grained and coarse-grained attention mechanisms to compose the MGAN framework. It also has an aspect alignment loss to depict the aspect-level interactions among aspects that have the same context.\\
\textbf{TNET}\cite{li2018transformation}. This system employs a CNN layer to extract salient features from the transformed word representations originated from a bi-directional RNN layer. Between the two layers, TNET has a component to generate target-specific representations of words while incorporating a mechanism for preserving the original contextual information from the RNN layer.\\
\textbf{BERT-DK}\cite{xu_bert2019}\footnote{\url{https://github.com/howardhsu/BERT-for-RRC-ABSA}}. This is the BERT-based model \cite{devlin2018bert}. It achieved the state-of-the-art results on ASC recently. 
Based on BERT, it first performs masked language modeling and then next sentence prediction on pre-trained BERT weights using domain (laptop or restaurant) reviews. Then it is fine-tuned using supervised ASC data.
We choose BERT-DK because of its easy-to-understand implementation without extra supervised tasks (such as reading comprehension) and its performance.
We further challenge this model by removing the aspects from the test example as there is no architecture change in doing so. In this way, the BERT-DK model has no way to check the aspect during testing. We want to see whether its performance on the \emph{Full Test Set} is affected much or not. Note that this is not a traditional sentence-level classifier as the training process is still under ASC task. 

\begin{table}[t]
\centering
\scalebox{0.66}{
    \begin{tabular}{l|c c|c c}
    \hline
     & \bf{Laptop} & & \bf{Rest.} & \\
    \hline
    






     & Acc. & MF1 & Acc. & MF1 \\
    \hline
    RAM\cite{chen2017recurrent} & & & & \\
    on Full Test Set & 74.49 & 71.35 & 80.23 & 70.8 \\
    on Contrastive Test Set & 41.87 & 38.65 & 52.19 & 55.19 \\     
    \hline
    AOA\cite{huang2018aspect} & & & & \\
    on Full Test Set & 74.5 & - & 81.2 & - \\
    on Contrastive Test Set & 42.86 & 33.53 & 42.98 & 33.66 \\
    \hline
    MGAN\cite{li2018exploiting} & & & & \\
    on Full Test Set & 75.39 & 72.47 & 81.25 & 71.94 \\
    on Contrastive Test Set & 46.8 & 43.38 & 53.95 & 57.64 \\
    \hline
    TNET\cite{li2018transformation} & & & & \\
    on Full Test Set & 76.54 & 71.75 & 80.69 & 71.27 \\
    on Contrastive Test Set & 49.75 & 49.86 & 56.58 & 58.05 \\
    \hline
    BERT-DK\cite{xu_bert2019} & & & & \\
    on Full Test Set & 76.9 & 73.65 & 84.21 & 76.2 \\
    on Full Test Set \textbf{w/o} aspect & \underline{76.0} & \underline{73.05} & \underline{80.03} & \underline{72.95} \\
    on Contrastive Test Set & 51.13 & 50.04 & 65.53 & 66.92 \\
    \hline
    \hline
    BERT-DK & Acc. & MF1 & Acc. & MF1 \\
    \hline
    + Manual Re-weighting & & & & \\
    on Full Test Set & 75.41 & 71.99 & 84.36 & 76.35 \\
    on Contrastive Test Set & 53.45 & 52.76 & 68.03 & 69.51 \\
    \hline
    + Focal Loss\cite{lin2017focal} & & & & \\
    on Full Test Set & 76.33 & 73.24 & 84.57 & 76.56 \\
    on Contrastive Test Set & 51.48 & 50.43 & 66.4 & 67.14 \\
    \hline
    + ARW & & & & \\
    on Full Test Set & 73.71 & 69.63 & 84.5 & 77.58 \\
    on Contrastive Test Set & 57.29 & 56.53 & 73.99 & 74.63 \\
    \hline
    + ARW w/ manual initial weighting & & & & \\
    on Full Test Set & 70.08 & 65.89 & 84.48 & 77.41 \\
    on Contrastive Test Set & 55.37 & 54.68 & \textbf{75.31} & \textbf{75.81} \\
    \hline
    + ARW \textbf{w/o} $\text{Contra}(\cdot)$ & & & & \\
    on Full Test Set & \textbf{77.23} & \textbf{73.81} & \textbf{85.35} & \textbf{78.46} \\
    on Contrastive Test Set & \textbf{61.08} & \textbf{60.34} & \textbf{71.84} & \textbf{72.66} \\
    \hline
    \end{tabular}
}
\caption{Performance of ASC baselines and the proposed ARW Scheme on both \emph{Full Test Set} and \emph{Contrastive Test Set}; the BERT-DK model is further tested on examples by removing aspects as in (\emph{on Full Test Set w/o aspect}).}
\label{tbl:failure}
\end{table}

\subsubsection{Baseline Result Analysis}
From Table \ref{tbl:failure}, we can see those existing ASC classifiers perform poorly on the contrastive test sets, which contain real ASC examples only.
To answer RQ1, we find that all baselines have significant drops on both Accuracy (Acc.) and F1 score as most existing models reach more than 70\% on both accuracy and F1 on the full test set.
Lastly, when the aspects are dropped from the input (\emph{on Full Test Set w/o aspect}), the BERT-DK ASC classifier dropped a little and still comparable to other baselines on the full test set. Since this experiment has no access to aspects but just the review sentences, it indicates that the model \textbf{DOES NOT} count on aspects much in doing aspect-level sentiment classification. 

\subsection{ARW}
The results of the above subsection justify the need for evaluating ASC on the contrastive test set and the need to improve the performance on that set. Since an ideal ASC should also be fully functional on none contrastive sentences, we still need to evaluate ARW and baselines on the full test set. In this set of experiments, we focus on ARW alone with various re-weighting schemes. 
\subsubsection{Compared Methods} We use BERT-DK as a base model to compare the following re-weighting schemes.\\
\textbf{+Manual Re-weighting}
This baseline uses pre-defined weights for examples from contrastive / non-contrastive sentences.
To answer RQ2, a natural way to balance the examples from contrastive sentences and non-contrastive sentences is to use the number of examples as weights.
To do so, we count the number of training examples $C_c$ from contrastive sentences and give them weights $(n-{C_c})$ and other examples weights $C_c$, where $n$ is the total number of training examples. So examples from contrastive sentences are expected to receive higher weights.
These weights are again re-normalized within a batch. Note that we also experimented with a number of other manual weighting schemes and this method does the best. \\
\textbf{+Focal Loss} To answer RQ4, we leverage the famous focal loss in object detection. The weight for each example is computed as $(1-p)^\gamma$, where $p$ is the probability of prediction on the ground-truth label (from softmax) and $\gamma$ is a hyper-parameter. We search this hyper-parameter and use $\gamma = 2.0 $ for results.\\ 
\textbf{+ARW} This is the proposed training scheme, which is intended to answer RQ3.\\ 
\textbf{+ARW w/ manual initial weighting} We further investigate the use of +Manual Re-weighting's weighting function as the initial weights and then use ARW for adaptive re-weighting.\\
\textbf{+ARW w/o $\text{Contra}(\cdot)$} This is the proposed training scheme without accessing the preprocessed labels for contrastive sentences, which is intended to answer RQ5. Note that this method discovers all incorrect examples, which may include examples from contrastive sentences. We search $\epsilon \in \{-0.2, -0.1, -0.05, 0.0, 0.05, 0.1, 0.2\}$ and use $\epsilon = -0.05 $ for results.


\subsubsection{Hyper-parameters}
For all methods, we use Adam optimizer and set the learning rate to 3e-5. The batch size is set as 32.
To perform model selection, we hold out 150 examples from the training set as a validation set.
We experimentally found that ARW takes longer time to converge compared with the ordinary training of a BERT-based model. For the \emph{Laptop} domain, it typically converges on the 8th or 9th epoch; for the \emph{restaurant} domain it converges on the 5th or 6th epoch. So we set the maximum epochs to 12.
All results are averaged over 10 runs.

\subsubsection{ARW Result Analysis}

\begin{table}[t]
\centering
\scalebox{0.7}{
    \begin{tabular}{l|c|c}
    \hline
    & \bf{Laptop} & \bf{Restaurant} \\
    \hline
    \# total examples & 2163 & 3452 \\
    \hline
    BERT-DK & & \\
    \# incorrect examples from contra. sent. & 148 & 228 \\
    \hline
    BERT-DK +ARW w/o $\text{Contra}(\cdot)$ & & \\
    \# incorrect examples from contra. sent. & \textbf{47} & \textbf{201} \\
    \hline
    \end{tabular}
}
\caption{Number of incorrectly predicted training examples (\# incorrect examples from contra. sent.) from contrastive sentences in one run of training: the training of the model is early stopped by validation set.}
\label{tbl:pred}
\end{table}

The results are also shown in Table \ref{tbl:failure}.
To answer RQ2, we observe that manual re-weighting improves the performance on laptop and restaurant about 3\% for the contrastive test sets.
After manual re-weighting, the performance on the full test set improves on restaurant but drops on laptop slightly. The reason could be that manual weights are not perfect for learning, which may overemphasize rare examples from contrastive sentences in the laptop training data.

To answer RQ3 and RQ5, we find that \emph{BERT-DK + ARW w/o $\text{Contra}(\cdot)$} mostly outperforms other baselines. If we compare with \emph{BERT-DK}, it is around 10\% of improvement for laptop and 6\% for restaurant.
Regarding the overall performance on the full test set, \emph{BERT-DK + ARW w/o $\text{Contra}(\cdot)$} has a marked improvement overall in the restaurant domain.
When we examine the examples, the contribution is largely from neutral examples.
Its performance on laptop is slightly better than \emph{BERT-DK}.
One reason could be that the examples from contrastive sentences are too rare compared to annotation errors in laptop. So the model learns some annotation errors.
Overall, these numbers indicate that \emph{BERT-DK + ARW w/o $\text{Contra}(\cdot)$} still functions well overall based on the traditional evaluation of ASC, but significantly improves the performance on contrastive sentences which truly test the aspect-level sentiment classification ability.
Further, we notice that both \emph{BERT-DK + ARW} and \emph{BERT-DK + ARW w/ manual initial weighting} tends to overfit the examples from contrastive sentences. It dropped a lot on sentences with singular polarity for laptop. For restaurant, \emph{BERT-DK + ARW w/ manual initial weighting} has the best performance on the contrastive test set, indicating manual re-weighting yields better weights initialization than uniform weights initialization in \emph{BERT-DK + ARW}.

To answer RQ4, we notice that focal loss does not perform very well for our problem.
Its performance on contrastive set is slightly better than BERT-DK. We believe the reason is that the numeric number of probability cannot explicitly distinguish whether the model is making a mistake on one example or not and thus provide poor weight to examples from contrastive sentences.

To answer RQ6,
we further investigate the behavior of both BERT-DK and BERT-DK + ARW w/o $\text{Contra}(\cdot)$ when their training is early stopped by the validation set, as shown in Table \ref{tbl:pred}.
We notice that the normal training of deep learning model (BERT-DK) naturally leaves more examples from contrastive sentences unresolved, justifying the reason why BERT-DK has poor performance on contrastive test set.
BERT-DK + ARW w/o $\text{Contra}(\cdot)$ obviously takes care of more examples from contrastive sentences before validation set finds the best model.

\subsubsection{Error Analysis}
Regarding errors for the \emph{Contrastive Test Set}, we noticed that given the limited number of contrastive sentences in training, some implicit sentiment transitions (or switching, such as no word like ``but'', etc.) 
is hard to learn (e.g., ``The screen is great and I can live with the keyboard's slightly smaller size.''). Also, contrastive sentences with \emph{neutral} polarity may be harder. This is because there may be no transition, but just one aspect with \emph{pos}/\emph{neg} opinion and one aspect with no opinion (\emph{neutral}). We believe using larger unlabeled corpora for training could benefit the contrastive test set. We leave that to our future work.
For the \emph{Full Test Set}, diverse and rare opinion expressions are also a very challenging problem to solve. Further, some fine-grained or uncommon opinion expressions are even hard to recognize by human annotators, resulting in annotation errors.

\section{Related Work}
\label{sec:rel}
Aspect sentiment classification (ASC) \cite{hu2004mining} is an important task in sentiment analysis~\cite{pang2002thumbs,liu2015sentiment}. It is different from document or sentence-level sentiment classification (SSC) \cite{pang2002thumbs,kim2014convolutional,he2011self,he2011automatically} as it focuses on fine-grained opinion on each specific aspect \cite{shu-etal-2017-lifelong,xu-etal-2018-double}. It is either studied as a single task or a joint learning task together with aspect extraction \cite{wang2017coupled,li2017deep,li2018unified}. The problem has been widely dealt with using neural networks in recent years \cite{dong2014adaptive,nguyen-shirai-2015-phrasernn,li2018transformation}.
Memory network and attention mechanisms are extensively applied to ASC, e.g., \cite{tang2016aspect,wang2016recursive,wang2016attention,ma2017interactive,chen2017recurrent,ma2017interactive,tay2018learning,he2018effective,liu2018content,xu2019review}. Memory networks \cite{weston2014memory,sukhbaatar2015end} are a type of neural networks that typically require two inputs and learn to have interactions between those two inputs via attention mechanisms~\cite{bahdanau2014neural}.
ASC is also studied in transfer learning or domain adaptation settings, such as leveraging large-scale corpora that are unlabeled or weakly labeled (e.g., using overall rating of a review as the label) \cite{xu_bert2019,he-EtAl:2018,xu2018lifelong} and transferring from other tasks/domains \cite{li2018exploiting,wang2018lifelong,wang2018target}. 

Contrastive opinions are studied as a topic modeling problem in \cite{ibeke-etal-2017-extracting} to discover constrastive opinions on the same opinion target from different holders, as in discussions. However, to the best of our knowledge, existing approaches and evaluations do not focus on contrastive sentences in aspect-based sentiment classification that having opposite opinions on different aspects from the same opinion holder. But those sentences or opinions truly reveal the capability of ASC models. 
 
The rare instance problem can be regarded as an imbalanced data problem in machine learning in general. Most existing studies in machine learning on imbalanced data focus on imbalanced classes or skewed class distributions, e.g., some classes with very few examples  \cite{huang2016learning,buda2018systematic,tantithamthavorn2018impact,johnson2019survey}. 
Object detection is a popular problem \cite{shrivastava2016training,lin2017focal} in computer vision for detecting long-tailed and imbalanced classes of examples given it is almost impossible to manually rebalance objects appear within an image.
In \cite{lin2017focal}, loss-based weights are proposed to automatically adjust weights without explicitly re-balance the complex class distribution. 
 
 Our example re-weighting algorithm is related to AdaBoost~\cite{freund1997decision}, which is a well-known ensemble algorithm that makes predictions collectively via a sequence of weak classifiers. When building each weak classifier, the weak learner tries to focus on the examples that are classified wrongly by the previous classifier. Weighted voting of these weak classifiers is used as the final ensemble classifier. Our work is different as we don’t build a sequence of classifiers as in AdaBoost but only one classifier. Neither is our model an ensemble model. Our weight updating is also different from AdaBoost as we do it in each epoch of training.
 However, our approach is in a similar spirit to that in AdaBoost on how to discover the weakness of the existing model on the training set. But we aim to improve the training process of a deep learning model by adaptively discovering incorrect examples that cover contrastive sentences and give them higher weights to focus on for subsequent training process. We also notice that AdaBoost is not frequently used in deep learning~\cite{schwenk2000boosting,mosca2017deep} probably due to the complexity of deep learning models which are not weak learners.
 
 Example (or instance) (re-)weighting is also leveraged in transfer learning and domain adaptation \cite{jiang-zhai-2007-instance,foster2010discriminative,xia2013instance,wang2017instance} and sentiment analysis \cite{pappas2014explaining}, but the purpose of weighting and weighting methods are entirely different. Re-weighting is commonly used to deal with noises in the training data as well. However, its focus is to weight down possible noisy training examples/instances~\cite{rebbapragada2007class}. It is not used to improve the hard but critical examples during training like what we do.

\section{Conclusion}
In this work, we observed a key failure of existing ASC classifiers. That is, they have great difficulty to classify contrastive sentences with multiple aspects and multiple different opinions, which are, in fact, the true test of aspect sentiment classifiers. We further showed that this difficulty is mainly caused by the fact that contrastive sentences are rare. One solution to this problem is to assign higher weights to such examples during training. However, instead of going for the solution that assigns higher weights manually, we proposed an automatic adaptive method ARW that discovers those incorrect examples from contrastive sentences during a certain stage of the training and adaptively assign them higher weights to improve their training. Experimental results show that our method is highly effective in handling contrastive sentences that are crucial for the ASC task and at the same time it also works very well on the full test set.

\bibliography{emnlp-ijcnlp-2019}
\bibliographystyle{acl_natbib}

\end{document}